\documentclass[letterpaper, 10 pt, conference]{ieeeconf}  

\IEEEoverridecommandlockouts                              

\usepackage{multirow} 
\usepackage{graphicx} 
\usepackage{svg} 
\usepackage{amsmath} 
\usepackage{amsfonts} 
\usepackage{booktabs} 
\usepackage{cite} 
\usepackage{makecell} 
\usepackage{url} 
\usepackage{caption} 
\usepackage{mathrsfs} 

\newcommand{\norm}[1]{\left\lVert #1 \right\rVert}

\title{\LARGE \bf
Orbeez-SLAM: A Real-time Monocular Visual SLAM\\ with ORB Features and NeRF-realized Mapping
}

\author{Chi-Ming Chung$^{1}$, Yang-Che Tseng$^{1}$, Ya-Ching Hsu$^{1}$, Xiang-Qian Shi$^{1}$, Yun-Hung Hua$^{1}$, Jia-Fong Yeh$^{1}$, 
\\ Wen-Chin Chen$^{1}$, Yi-Ting Chen$^{2}$ and Winston H. Hsu$^{1, 3}$
\thanks{$^{1}$National Taiwan University}%
\thanks{$^{2}$National Yang Ming Chiao Tung University}%
\thanks{$^{3}$Mobile Drive Technology}%
}

\begin{document}



\twocolumn[{
\renewcommand\twocolumn[1][]{#1}
\maketitle
\thispagestyle{empty}
\pagestyle{empty}
\begin{center}
    \captionsetup{type=figure}
    \includegraphics[width=.9\linewidth]{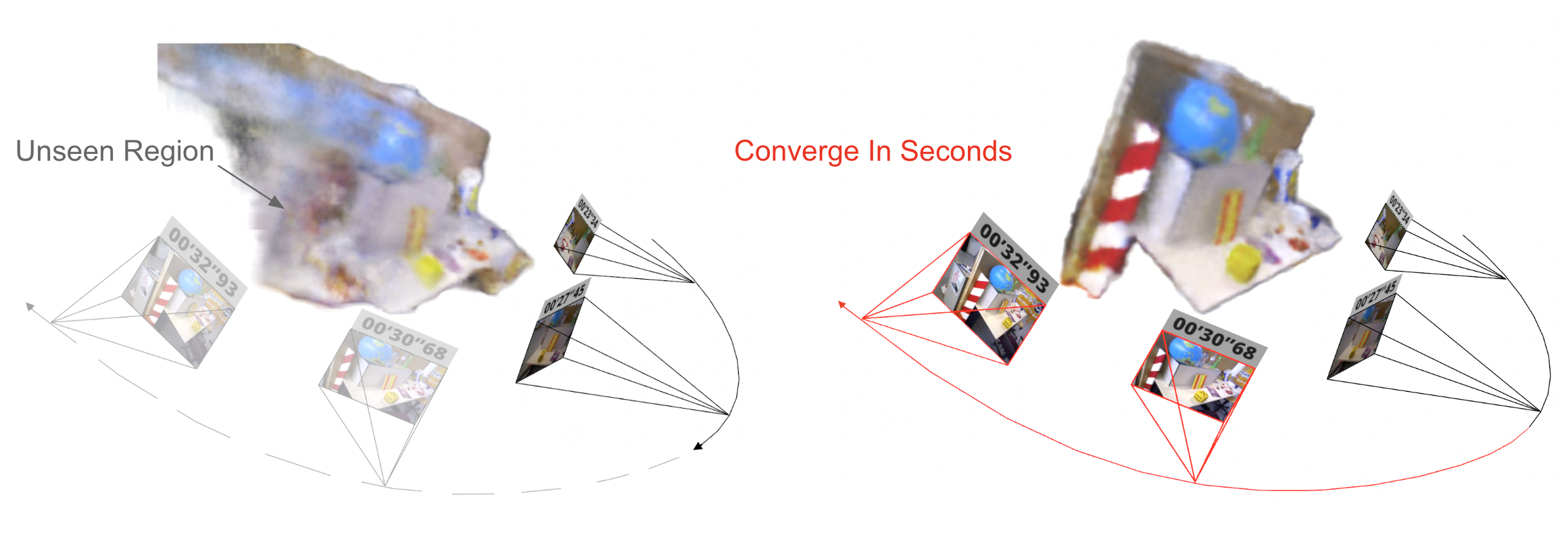}
    \captionof{figure}{\textbf{Orbeez-SLAM process.} The numbers above the camera represent the timestamp of tracking. The left image shows NeRF result at 27 seconds training from scratch. The right image shows that after seeing the unseen region on the left (i.e., novel views), our NeRF model can update the region in seconds. Our Orbeez-SLAM is real-time and pre-training-free. }
    \label{fig:overview}
\end{center}
}]
{
  \footnotetext[1]{National Taiwan University}
  \footnotetext[2]{National Yang Ming Chiao Tung University}
  \footnotetext[3]{Mobile Drive Technology}
}

\begin{abstract}
A spatial AI that can perform complex tasks through visual signals and cooperate with humans is highly anticipated. To achieve this, we need a visual SLAM that easily adapts to new scenes without pre-training and generates dense maps for downstream tasks in real-time. None of the previous learning-based and non-learning-based visual SLAMs satisfy all needs due to the intrinsic limitations of their components. In this work, we develop a visual SLAM named Orbeez-SLAM, which successfully collaborates with implicit neural representation and visual odometry to achieve our goals. Moreover, Orbeez-SLAM can work with the monocular camera since it only needs RGB inputs, making it widely applicable to the real world. Results show that our SLAM is up to 800x faster than the strong baseline with superior rendering outcomes. Code link: \url{https://github.com/MarvinChung/Orbeez-SLAM}.

\end{abstract}

\section{Introduction}
An intelligent spatial AI that can receive visual signals (RGB-D images) and cooperate with humans to solve complicated tasks is highly valued. To efficiently understand semantics knowledge from the environment and act like a human, spatial AI requires a core component named visual simultaneous localization and mapping (SLAM). The visual SLAM should quickly adapt to new scenes without pre-training and generate real-time fine-grained maps for downstream tasks, such as domestic robots. However, traditional visual SLAMs \cite{murORB2,engel14eccv} mainly focus on localization accuracy and only provide crude maps. To this end, this work aims to develop a visual SLAM with the aforementioned properties.

 To compute dense maps, a recent learning-based visual SLAM, Tandem \cite{koestler2021tandem}, leverages the truncated signed distance function (TSDF) fusion to provide a dense 3D map. As claimed in \cite{koestler2021tandem}, Tandem achieves real-time inference and can work with a monocular camera. However, depth estimation is involved in the TSDF fusion, and the depth estimation module in Tandem needs pre-training before inference, which limits its adaptability to a novel scene significantly different from pre-trained scenes. 
 
 Neural Radiance Field (NeRF) \cite{mildenhall2020nerf}, another implicit neural representation, does not require depth supervision during training and can be trained from scratch at the target scene. Due to this attribute, using NeRF as the map in visual SLAM is a potential direction. Two latest NeRF-SLAMs \cite{Sucar:etal:ICCV2021, Zhu2022CVPR} echo our motivations. Among them, iMAP \cite{Sucar:etal:ICCV2021} is the first work that lets NeRF serve as the map representation in SLAM. Meanwhile, it optimizes the camera pose via back-propagation from NeRF photometric loss. Then, NICE-SLAM \cite{Zhu2022CVPR} extends iMAP and develops a hierarchical feature grid module. The module allows NICE-SLAM to scale up for large scenes without catastrophic forgetting. Nevertheless, the above NeRF-SLAMs need RGB-D inputs since they optimize camera pose purely through the neural network without visual odometry (VO), causing bad initial localizations. In other words, they still need depth information to guide the 3D geometry. Besides, a notable shortcoming of NeRF is its slow convergence speed. Specifically, it utilizes lots of rendering, which makes real-time training NeRF floundering. By observing this, instant-ngp \cite{mueller2022instant} compensates for the training speed issue. With the help of the multi-resolution hash encoding and the CUDA framework \cite{tiny-cuda-nn}, instant-ngp can train NeRFs in a few seconds. 

 To tackle the above shortcomings, we seek to develop a monocular visual SLAM that is pre-training-free and achieves real-time inference for practical applications. To this end, we propose Orbeez-SLAM, combining feature-based SLAM (e.g., ORB-SLAM2 \cite{murORB2}) and a NeRF based on the instant-ngp framework \cite{mueller2022instant}. Different from \cite{Sucar:etal:ICCV2021, Zhu2022CVPR}, we emphasize that VO (in ORB-SLAM2) can provide a better camera pose estimation even at the early stage of the training, which lets Orbeez-SLAM can work with monocular cameras, i.e., without depth supervision. Moreover, we simultaneously estimate camera poses via VO and update the NeRF network. Notably, the training process is online and real-time without pre-training, as depicted in Fig. \ref{fig:overview}. As a result, Orbeez-SLAM can render dense information such as the depth and color of scenes. Besides, it is validated in various indoor scenes and outperforms NeRF-SLAM baselines on speed, camera tracking, and reconstruction aspects. To summarize, our contributions are threefold:

\begin{itemize}
    \item We propose Orbeez-SLAM, the first real-time monocular visual SLAM that is pre-training-free and provides dense maps, tailored for spatial AI applications.
    \item By combining visual odometry and a fast NeRF framework, our method reaches real-time inference and produces dense maps.
    \item We extensively validate Orbeez-SLAM with state-of-the-art (SOTA) baselines on challenging benchmarks, showing superior quantitative and qualitative results.
\end{itemize}

\section{Related works}
\subsection{Implicit neural representations}
To represent a 3D scene, explicit representations (e.g., point clouds) need huge space to store information. By contrast, implicit surface representations, such as signed distance functions (SDF), alleviate the space issue and have been widely developed in recent years. Among them, some works \cite{takikawa2021nglod, Park_2019_CVPR} leverage neural networks to learn the implicit function, called implicit neural representations (INRs). With the property of continuous representation of the signals, INRs demonstrate several advantages: (a) they are not coupled to the spatial dimension/resolution of input signals, and (b) they can predict/synthesize the unobserved regions.

Besides, NeRF, a novel and popular INR, has illustrated its success in novel view synthesis \cite{mildenhall2020nerf, martinbrualla2020nerfw, barron2021mipnerf, barron2022mipnerf360}. Nonetheless, most NeRFs assume that the camera pose is known. Thus, COLMAP \cite{schoenberger2016sfm, schoenberger2016mvs} is often used to estimate intrinsic and extrinsic (camera poses) in NeRF-related works. In addition, a few works \cite{wang2021nerfmm, SCNeRF2021, lin2021barf} optimize camera poses via NeRF photometric loss, but the process requires a long training time. Hence, as aforementioned, instant-ngp \cite{mueller2022instant} develops a framework that can train NeRFs in a few seconds, leveraging the multi-resolution hash encoding and CUDA platform \cite{tiny-cuda-nn}. 

Intuitively, implicit surface representations can serve as maps in visual SLAM systems. For instance, some studies \cite{6162880, 8255617} leverage the SDF to construct the map. Besides, two recent NeRF-SLAMs  \cite{Sucar:etal:ICCV2021, Zhu2022CVPR} pave the way for cooperating NeRFs and visual SLAM. However, they need RGB-D inputs and show a slow convergence speed, which do not satisfy our needs.
Therefore, we aim to build a NeRF-SLAM to generate a dense map in real-time. Moreover, our work can work with the monocular camera and train from scratch at the target scene without a lengthy pre-training process.

\subsection{Visual SLAM systems}
Traditional visual SLAMs reveal strengths in outdoor and large scenes. Also, they can rapidly compute accurate locations but lack the fine-grained information from scenes.  There are two categories of visual SLAMs, feature-based and direct SLAM. Feature-based SLAMs \cite{murAcceptedTRO2015, murORB2, ORBSLAM3_TRO} extract and match image features between frames and then minimize the reprojection error. Besides, Direct SLAM \cite{engel14eccv} uses pixel intensities to localize and minimize the photometric error.

To satisfy the needs of spatial AI, we require a visual SLAM that provides a dense map for complicated tasks. Several works \cite{koestler2021tandem, Sucar:etal:ICCV2021, Zhu2022CVPR} achieve this objective under deep learning techniques.
However, they either need pre-training \cite{koestler2021tandem} that limits the adaptation capability or optimize the camera poses and network parameters only relying on NeRF photometric loss and depth supervision \cite{Sucar:etal:ICCV2021, Zhu2022CVPR}, lacking the knowledge of VO. Thus, we develop Orbeez-SLAM that eliminates these drawbacks by considering the VO guidance and fast NeRF implementation. Consequently, Orbeez-SLAM is pre-training-free for novel scenes and reaches real-time inference (with online training).





\section{Preliminaries}

\subsection{NeRF}
\label{sec:nerf}
NeRF \cite{mildenhall2020nerf} reconstructs a 3D scene by training on a sequence of 2D images from distinct viewpoints. A continuous scene can be represented as a function $F: (\mathbf{x}, \mathbf{d}) \rightarrow (\mathbf{c}, \sigma),$ where $\mathbf{x}$ is the 3D location, $\mathbf{d} = (dx, dy, dz)$ is the 3D Cartesian unit vector representing the direction $(\theta, \phi)$. Outputs are color $\mathbf{c} = (r,g,b)$ and volume density $\sigma$. Such a continuous representation, as a function, can be approximated with an MLP network $F(\Theta)$ by optimizing the weight $\Theta$.

Following above definitions, for any ray $\mathbf{r}(t) = \mathbf{o}+t\mathbf{d}$, the color $C(\mathbf{r})$ within the near and far bounds $[t_n, t_f]$ can be obtained through the intergral of products of transmittance $T(t)$, volume density and the color at each point, i.e.,
\begin{equation}
\label{eq:colorformula}
C(\mathbf{r}) = \int_{t_n}^{t_f}T(t)\sigma(\mathbf{r}(t))\mathbf{c}(\mathbf{r}(t), \mathbf{d})dt,
\end{equation}
 where 
\begin{equation}
    T(t) = \exp\bigg(-\int_{t_n}^{t}\sigma(\mathbf{r}(s))ds\bigg).
\end{equation}

To feed the input rays into the neural network, $N$ discrete points $t_i \in [t_n, t_f], i \in [1, N]$ are sampled to estimate the color (Eq. \ref{eq:colorformula}) with the quadrature rule \cite{468400}:
\begin{equation}
    \label{eq:colorformula_discrete}
    \hat{C}(\mathbf{r})=\sum_{i=1}^{N}T_i(1-\exp(-\sigma_i(t_{i+1}-t_i)))\bold{c}_i,
\end{equation}
where 
\begin{equation}
    T_i = \exp\bigg(-\sum_{j=1}^{i-1}\sigma_j(t_{j+1}-t_j)\bigg).
\end{equation}
The weight of a sample point is denoted as
\begin{equation}
    \label{eq:weight}
    w_i = T_i(1-\exp(-\sigma_i(t_{i+1}-t_i)).
\end{equation}

\begin{figure}[!tp]
  \centering
  \includegraphics[width=\linewidth]{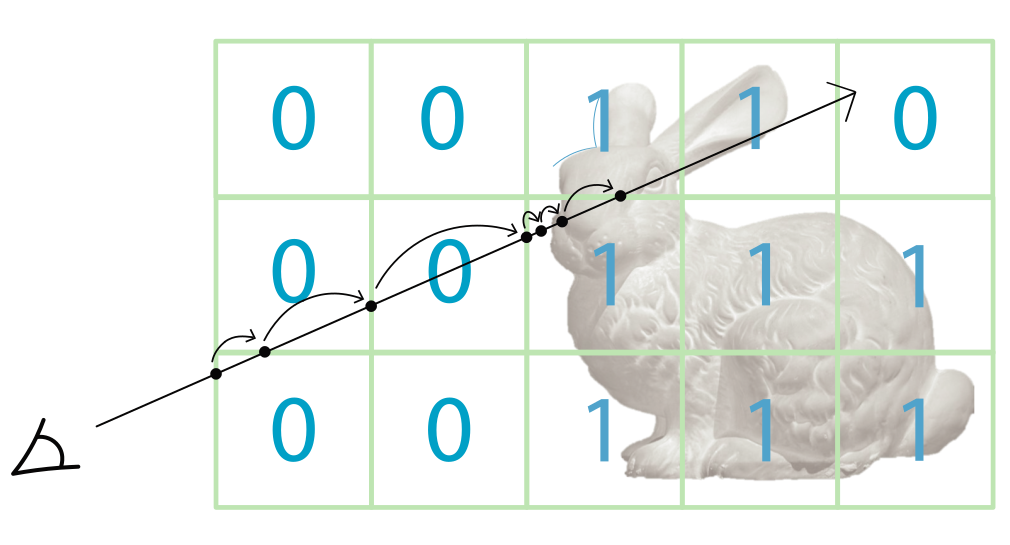}
  \caption{\textbf{Skip voxel strategy.} When sampling positions along a cast ray, a voxel is skipped if it is unoccupied (mark as 0); we sample voxels which intersect the surface (mark as 1).}
  \label{fig:AccelerateStructure}
\end{figure}

\subsection{Density grid}
The rendering equation (\ref{eq:colorformula}) in NeRF requires sampling positions on the ray. We should be only interested in the positions which intersect the surface since they contribute more to (\ref{eq:colorformula}). Some studies \cite{mildenhall2020nerf, barron2021mipnerf, barron2022mipnerf360} leverage a coarse-to-fine strategy that samples uniformly on the ray to find the density distribution via querying NeRF. After knowing the density distribution of the ray, they only sample those positions near the surface. However, these steps require frequent NeRF querying, which is time-consuming.

To tackle this, recent works \cite{mueller2022instant, liu2020neural, yu_and_fridovichkeil2021plenoxels, SunSC22, Clark_2022_CVPR} store the query results in density grid, and then the skip voxel strategy is usually applied, as shown in Fig. \ref{fig:AccelerateStructure}. In this work, we further extend the skip voxel strategy with knowledge of NeRF to process the ray-casting triangulation (see Section \ref{sec:triangulation}).


\section{Methodology}
\begin{figure*}[!tp]
  \centering
  \includegraphics[width=\textwidth]{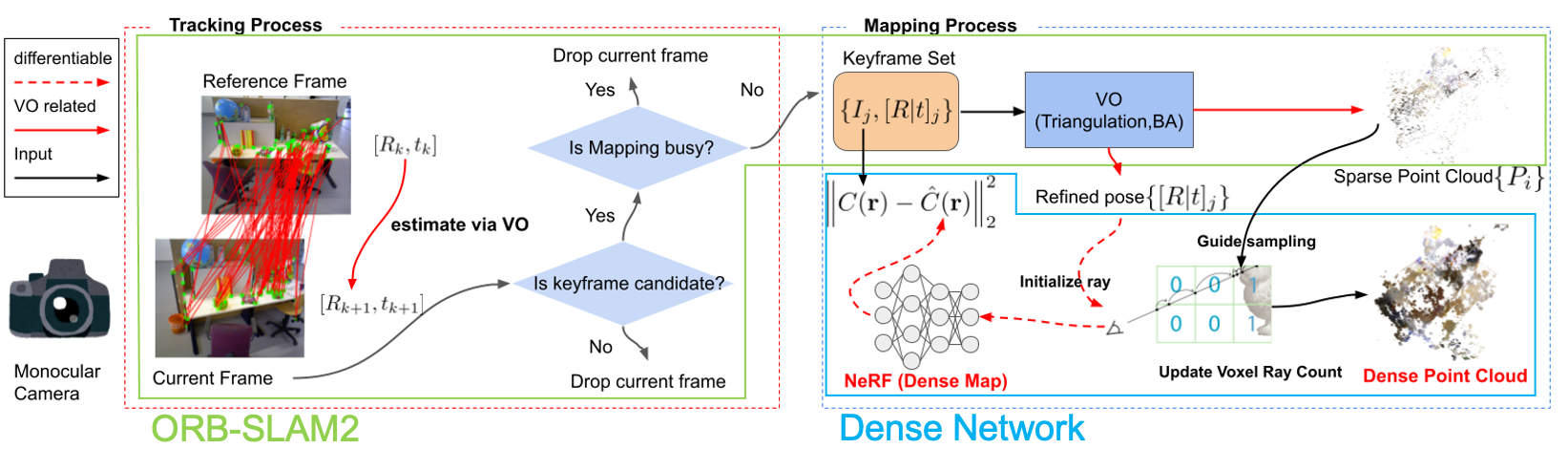}
  \caption{\textbf{System Pipeline.} The tracking and mapping processes run concurrently. A frame from the image stream must satisfy the two conditions to become a keyframe. The first condition filters out those frames with weak tracking results. The second condition drops the frame if the mapping process is busy. The tracking process provides camera poses estimation. The mapping process refines the camera poses and maintains the maps. We also show the dense point cloud generated from our proposed ray-casting triangulation which is introduced in Section  \ref{sec:triangulation}.}
  \label{fig:pipeline}
\end{figure*}
Unlike previous NeRF-SLAMs \cite{Sucar:etal:ICCV2021, Zhu2022CVPR} which require depth information to perceive geometry better, we develop Orbeez-SLAM that leverages VO for accurate pose estimations to generate a dense map with a monocular camera. Besides, it achieves pre-training-free adaptation and real-time inference. Next, the system overview is depicted in Sec. \ref{sec:overview} and the optimization objectives are described in Sec. \ref{section:Optimization}. At last, ray-casting triangulation is introduced in Sec. \ref{sec:triangulation}.

\subsection{System overview}
\label{sec:overview}
Fig. \ref{fig:pipeline} shows our system pipeline. The tracking process extracts the image features from the input image stream $I$ and estimates the camera poses via VO. The mapping system generates map points with triangulation and optimizes camera poses and map points with bundle adjustment (reprojection error). These map points represent a sparse point cloud. We then utilize the updated camera poses and the map to train NeRF. Since the process is differentiable, we can still optimize the camera poses from NeRF photometric loss. In the end, the NeRF can generate a dense map for downstream tasks. Moreover, this pipeline should work for any SLAM that provides sparse point cloud.

\begin{figure}[!bp]
  \centering
  \includegraphics[width=\linewidth]{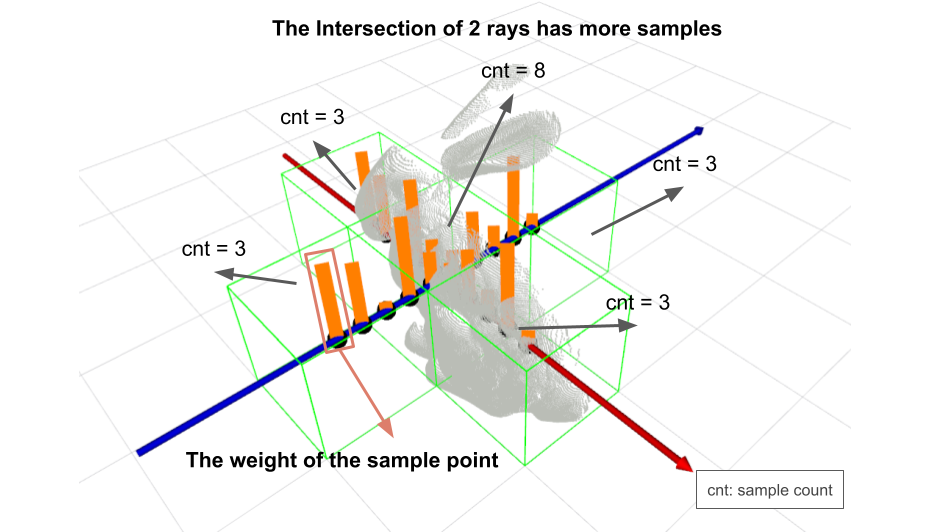}
  \caption{\textbf{Ray-casting triangulation in NeRF.} We record the sample count for each density grid voxel. If the weight of a voxel (See (\ref{eq:weight}) in Sec. \ref{sec:nerf}) exceeds the threshold to be a surface candidate, we add 1 to the voxel counter. Those voxels with high sample counts will likely contain surface and be added as map points for the dense point cloud.}
  \label{fig:triangulation}
\end{figure}

\subsection{Optimization}
\label{section:Optimization}
The following objectives are used to optimize Orbeez-SLAM: (a) pose estimation, (b) bundle adjustment, and (c) NeRF regression. Among them, (a) is in the tracking process, and (b) and (c) are conducted in the mapping process.


(a) \textit{Pose estimation:}
Reprojection error \cite{DBLP:journals/corr/abs-1912-03858} is widely used in feature-based SLAM \cite{murAcceptedTRO2015, murORB2, ORBSLAM3_TRO} to estimate the pose, and its formulation is as follows: 
\begin{equation}
\label{eq:reprojection}
     L_{rpj} = \sum_{ij} \norm{u_{ij} - \pi(\mathscr{C}_j, P_i)}_{2}
\end{equation}
where $u_{ij}$ is the pixel position on the image, which is observed by the $j$th camera $\mathscr{C}_j$ and is projected by the $i$th 3D point. The $\pi(\mathscr{C}_{j}, P_{i})$ projects the 3D map point $P_{i}$ to the pixel coordinate via $\frac{1}{Z}K_{j}(R_{j}P_{i}+t_{j})$,  where $P_{i} = [X,Y,Z]^{T}$ and $K_{j}, [R|t]_{j}$ are the intrinsic and extrinsic (world to camera) described by $\mathscr{C}_{j}$.  We optimizes the camera poses $\{[R|t]_{j}\}$ by minimizing the reprojection error:
\begin{equation}
    \min \limits_{\{[R|t]_{j}\}} L_{rpj}
\end{equation}

(b) \textit{Bundle adjustment:}
After the triangulation step in VO, new map points are added to the local map. The bundle adjustment objective also minimizes the reprojection error for both the map point positions and the camera poses: 

\begin{equation}
\label{eq:BA}
     \min \limits_{\{[R|t]_{j}\},\{P_{i}\}} L_{rpj}
\end{equation}

Minimizing (\ref{eq:reprojection}) is actually a nonlinear least square problem. We solve these two objectives by the Levenberg-Marquardt method, followed \cite{murORB2}. The bundle adjustment optimizes the camera poses of keyframes $(K,[R|t])$ and observable map points in these keyframes. Then, these optimized keyframes and map points are passed to the NeRF.

(c) \textit{NeRF regression:}
NeRF minimizes the photometric error by regressing the image color. A ray can be formulated by giving a keyframe $(K,[R|t])$ and a pixel coordinate $[u,v]$: 
\begin{equation}
\label{eq:pose2ray}
r(d) = (-R^{T}t+dR^{T}K^{-1}[u,v,1]^{T}), d \in \mathbb{R}    
\end{equation}

By applying the \textbf{skip voxel} strategy mentioned in Fig. \ref{fig:AccelerateStructure}, we sample positions on the ray were near to the surface. Finally, the NeRF photometric loss is the L2 norm between predicted color $\hat{C}(\mathbf{r})$ and the pixel color ${C}(\mathbf{r})$.
\begin{equation}
\label{eq:color_diff}
L_{pht} = \sum_{ij} \norm{C(\mathbf{r}_{ij}) - \hat{C}(\mathbf{r}_{ij})}_{2}
\end{equation}
where $C(\mathbf{r}_{ij})$ is the observed color of ray $j$ in image $I_{i}$. Since (\ref{eq:pose2ray}) is differentiable, both camera extrinsic $\{[R|t]_{i}\}$ and network parameters $\Theta$ can be optimized by $L_{pht}$. But, after examinations (cf. Tab. \ref{tab:ablation}), we only optimize $\{[R|t]_{i}\}$ by (\ref{eq:reprojection}).

\subsection{Ray-casting triangulation}
\label{sec:triangulation}
In Fig. \ref{fig:AccelerateStructure}, we show that the density grid can accelerate the rendering process. However, this structure only considers a ray and highly relies on the density prediction of the NeRF model. We additionally store the number of sampling times for each voxel. A voxel that frequently blocks the casting ray is more likely to be the surface, as shown in Fig. \ref{fig:triangulation}. For noise rejection, we only triangulate points that lie within voxels that are scanned by rays frequently enough. We chose 64 as the threshold for practical implementation since such a value has the best visualization, according to our experience. We also utilize the data structure's map point generated from the sparse point cloud. Since the map point's surroundings are more likely to be the surface, we add a significant number to the sample counter of the density grid. We claim that this method can find a more reliable surface and online generates map points with a training NeRF. Map points generated by this method are not optimized in (\ref{eq:BA}). We show the dense point cloud generated by this method in Fig. \ref{fig:pipeline}.

\section{Experiments}
\renewcommand{\subsubsection}[1]{\noindent{\textbf{#1}}}


\subsection{Experimental setup}

\subsubsection{Datasets.} For a fair comparison, we conduct our experiments on three benchmarks, TUM RGB-D \cite{sturm12iros}, Replica \cite{replica19arxiv}, and ScanNet \cite{dai2017scannet}, which provide extensive images, depths, and camera trajectories and are widely used in previous works.

\subsubsection{Baselines.} 
We compare proposed Orbeez-SLAM with two categories of baselines, (a) learning-based SLAM: DI-Fusion\cite{huang2021difusion}, iMap\cite{Sucar:etal:ICCV2021}, iMap$^{*}$(re-implemented in \cite{Zhu2022CVPR}), and NICE-SLAM \cite{Zhu2022CVPR}. (b) traditional based SLAM: BAD-SLAM\cite{8954208}, Kintinuous\cite{whelan2012kintinuous}, and ORB-SLAM2\cite{murORB2}.

\subsubsection{Evaluation settings.} In practice, monocular SLAM works validate the effectiveness under depth version since they cannot estimate the correct scale of the scenes without knowing depth. Also,  all previous NeRF-SLAMs require depth supervision. Thus, all methods are verified on the depth version. We still demonstrate that Orbeez-SLAM can work with monocular cameras. Moreover, we extensively examine the efficacy from two aspects, tracking and mapping results.

\subsubsection{Metrics.} To evaluate the tracking results, we report the absolute trajectory error (ATE), which computes the root mean square error (RMSE) between the ground truth (GT) trajectory and the aligned estimated trajectory. For the mapping results, we extend PSNR and Depth L1 metrics that are often used in NeRF to NeRF-SLAM. PSNR assesses the distortion rate of NeRF rendered and GT images traversed by the GT trajectory. As for Depth L1, we calculate the L1 error of estimated and GT depth traversed by the GT trajectory.

Unlike in \cite{Sucar:etal:ICCV2021, Zhu2022CVPR}, we do not evaluate on meshes. We argue that assessing performance on meshes may be unfair because the mesh generation process by post processing NeRF is not unified. In addition, our setting has the following advantages:

\begin{itemize}
\item Numbers of sampled keyframes in distinct works are various while evaluating with GT trajectory provides a consistent standard.

\item The depth and PSNR can effectively reflect the quality of the geometry and radiance learned by NeRF.

\item Our setting can verify the methods on novel views since only keyframes (subset of GT trajectory) are used during training. Besides, even if the model backs up seen keyframes, the metric can still reveal that when they are localized to the wrong viewpoints. 
\end{itemize}

\begin{table}[!t]
  \centering
  \footnotesize
  \setlength{\tabcolsep}{0.7em}
  \vspace{2pt}
 \caption{\textbf{Tracking Results on TUM RGB-D.} ATE [\textbf{cm}] ($\downarrow$) is used. Learning-based and traditional SLAMs are separated by the middle line. Results of DI-Fusion, iMAP, iMAP$^{*}$, NICE-SLAM, BAD-SLAM and Kintinuous are from \cite{Zhu2022CVPR}. The best out of 5 runs are reported.  
    }

      \begin{tabular}{lccc}
      \toprule
         & \texttt{fr1/desk} &  \texttt{fr2/xyz} &  \texttt{fr3/office} \\
         \midrule
        {DI-Fusion\cite{huang2021difusion}} & 4.4 & 2.3 & 15.6 \\
        {iMAP}\cite{Sucar:etal:ICCV2021}      & 4.9 & 2.0 & 5.8  \\
        {iMAP$^*$\cite{Zhu2022CVPR}} & 7.2 & 2.1  & 9.0 \\
        
        {NICE-SLAM\cite{Zhu2022CVPR}} & 2.7 & 1.8 & 3.0 \\
        {Orbeez-SLAM (\textbf{Ours})} & 1.9 & \bf 0.3 & 1.0 \\
      \midrule
        {BAD-SLAM\cite{8954208}} & 1.7  & 1.1  & 1.7 \\
        {Kintinuous\cite{whelan2012kintinuous}} & 3.7  &  2.9  & 3.0 \\
        {ORB-SLAM2}\cite{murORB2} & \bf 1.6  & \bf 0.3  & \bf 0.9 \\
       \bottomrule
    \end{tabular}
    \vspace{2pt}
    \label{tab:tum_rmse}
\end{table}

\begin{table}[t!]
  \centering
  \footnotesize
  \setlength{\tabcolsep}{4pt}
  \caption{\textbf{Tracking Results on ScanNet.} ATE [\textbf{cm}] ($\downarrow$) is used. Results of iMAP$^{*}$, DI-Fusion and NICE-SLAM are from \cite{Zhu2022CVPR}.}
  \resizebox{\linewidth}{!}{
    \begin{tabular}{lccccccc}
      \toprule
         Scene ID & \multicolumn{1}{c}{\makecell{\texttt{0000}}} & \multicolumn{1}{c}{\makecell{\texttt{0059}}} &  \multicolumn{1}{c}{\makecell{\texttt{0106}}} & \multicolumn{1}{c}{\makecell{\texttt{0169}}} & \multicolumn{1}{c}{\makecell{\texttt{0181}}} & \multicolumn{1}{c}{\makecell{\texttt{0207}}} &
         Avg.\\
         \midrule
         {DI-Fusion ~\cite{huang2021difusion}} & 62.99 & 128.00 & 18.50 & 75.80 & 87.88 & 100.19 & 78.89 \\
        {iMAP$^*$~\cite{Sucar:etal:ICCV2021}} & 55.95 & 32.06 & 17.50 &70.51 & 32.10 & 11.91 & 36.67\\
        {NICE-SLAM\cite{Zhu2022CVPR}} & 8.64 & 12.25 & 8.09 & 10.28 & \textbf{12.93} & \textbf{5.59} & 9.63\\
        {Orbeez-SLAM (\textbf{Ours})} & \textbf{7.22} & 7.15 & \textbf{8.05} & \textbf{6.58} & 15.77 & 7.16 & \textbf{8.655} \\
        \midrule
        {ORB-SLAM2 \cite{murORB2}} & 7.57 & \textbf{6.92} & 8.30 & 6.90 & 16.42 & 8.78 & 9.15 \\
       \bottomrule
    \end{tabular}
    }
    \label{tab:scannet}
\end{table}

\begin{table}[!t]
\centering
\caption{\textbf{Reconstruction Results on Replica.} Depth L1 [\textbf{cm}] ($\downarrow$) and PSNR [\textbf{dB}] ($\uparrow$) are used. The values are averaged over office 0 to 4 and room 0 to 2. NICE-SLAM use GT depth during rendering color and depth. We show the results of NICE-SLAM w and w/o GT depth.}

\begin{tabular}{@{}lcccc@{}}
\toprule
\multicolumn{1}{c}{} & \begin{tabular}[c]{@{}c@{}}Depth $\downarrow$\\ \texttt{w/o GT}\end{tabular} & \begin{tabular}[c]{@{}c@{}}Depth $\downarrow$\\ \texttt{w/ GT} \end{tabular} & \begin{tabular}[c]{@{}c@{}}PSNR $\uparrow$\\ \texttt{w/o GT}\end{tabular}  & \begin{tabular}[c]{@{}c@{}}PSNR $\uparrow$\\ \texttt{w/ GT}\end{tabular} \\ 
\midrule
NICE-SLAM \cite{Zhu2022CVPR}  & 13.49 & \bf 4.22 & 17.74 & 24.60 \\
Orbeez-SLAM (\textbf{Ours}) & \bf 11.88 & \multicolumn{1}{c}{-} & \bf 29.25 & \multicolumn{1}{c}{-} \\ 
\bottomrule
\end{tabular}
\label{tab:replica_rec}
\end{table}

\subsubsection{Implementation Details.} We conduct all experiments on a desktop PC with an Intel i7-9700 CPU and a NVIDIA RTX 3090 GPU. We follow the official code in ORB-SLAM2 \footnote{\url{https://github.com/raulmur/ORB_SLAM2}} \cite{murORB2} and instant-ngp \footnote{\url{https://github.com/NVlabs/instant-ngp}} \cite{mueller2022instant} to implement Orbeez-SLAM. Note that Orbeez-SLAM inherits the loop-closing process from ORB-SLAM2 \cite{murORB2} to improve the trajectory accuracy. We do not cull the keyframe like ORB-SLAM2 to ensure the keyframe is not eliminated after passing to the NeRF.
The code is written in C++ and CUDA. About losses, reprojection error is optimized via g2o \cite{5979949} framework, and the photometric error in NeRF is optimized via tiny-cuda-nn framework \cite{tiny-cuda-nn}.  

\begin{table}[!t]
\centering
\vspace{2pt}
\caption{\textbf{Runtime Comparison.} Frame per second [\textbf{fps}] ($\uparrow$) when running on TUM RGB-D. We show that Orbeez-SLAM is much faster than the SOTA NeRF-SLAM. }
\begin{tabular}{cccc}
\hline
\multicolumn{1}{c}{} & fr1/desk      & fr2/xyz    & fr3/office    
\\ { \# images} & 613 & 3669 & 2585 
\\ \hline
NICE-SLAM \cite{Zhu2022CVPR}           & 0.056 & 0.028  & 0.037 \\
Orbeez-SLAM (\textbf{Ours})      & 19.210     & 22.725 & 21.542    \\ \hline\hline
\end{tabular}
\label{tab:speed}
\end{table}



\begin{table}[!t]
\centering

\caption{\textbf{Ablation study on Replica.} We demonstrate that optimizing camera poses only from reprojection error $L_{rpj}$ is better than from both $L_{rpj}$ and photometric error $L_{pht}$. }
\begin{tabular}{@{}ccccc@{}}
\toprule
\multicolumn{1}{c}{$L_{rpj}$} & \multicolumn{1}{c}{$L_{pht}$} & \multicolumn{1}{c}{ATE $\downarrow$} & \multicolumn{1}{c}{Depth $\downarrow$} & \multicolumn{1}{c}{PSNR $\uparrow$} \\ \midrule
\checkmark & \checkmark & 5.3 & 13.43 & 25.52 \\
\checkmark & & \bf 0.8 & \bf 11.88 & \bf 29.25\\ \bottomrule
\end{tabular}
\label{tab:ablation}
\end{table}

\subsection{Quantitative Results}
We aim to verify whether Orbeez-SLAM can produce accurate trajectories (ATE), precise 3D reconstructions (Depth), and detailed perception information (PSNR) under our challenging settings, i.e., real-time inference without pre-training. Previous works focus on the first two indicators. However, a dense map containing rich perception information is vital for spatial AI applications; thus, we also attend to this aspect.


\subsubsection{Evaluation on TUM RGB-D (small-scale) \cite{sturm12iros}.} 
TABLE \ref{tab:tum_rmse} lists tracking results of all methods. Note that Orbeez-SLAM outperforms all deep-learning baselines with a significant gap (top half). Besides, ORB-SLAM2 is our upper bound on the tracking results since our method is built on it. Nevertheless, Orbeez-SLAM only shows a minor performance drop while it provides a dense map generated by NeRF.

\subsubsection{Evaluation on ScanNet (large-scale) \cite{dai2017scannet}. }
As revealed in TABLE \ref{tab:scannet},  we obtain the best average results across all scenes. We assume the performance difference between us and ORB-SLAM2 is due to randomness. In addition, NICE-SLAM performs best in some cases, echoing the claimed strength for scaling-up scenes in \cite{Zhu2022CVPR}. Especially scenes 0181 and 0207 contain compartments. Improving performance in large scenes with rooms is one of future works.

\begin{figure*}[!t]
    \centering
    \includegraphics[width=\linewidth]{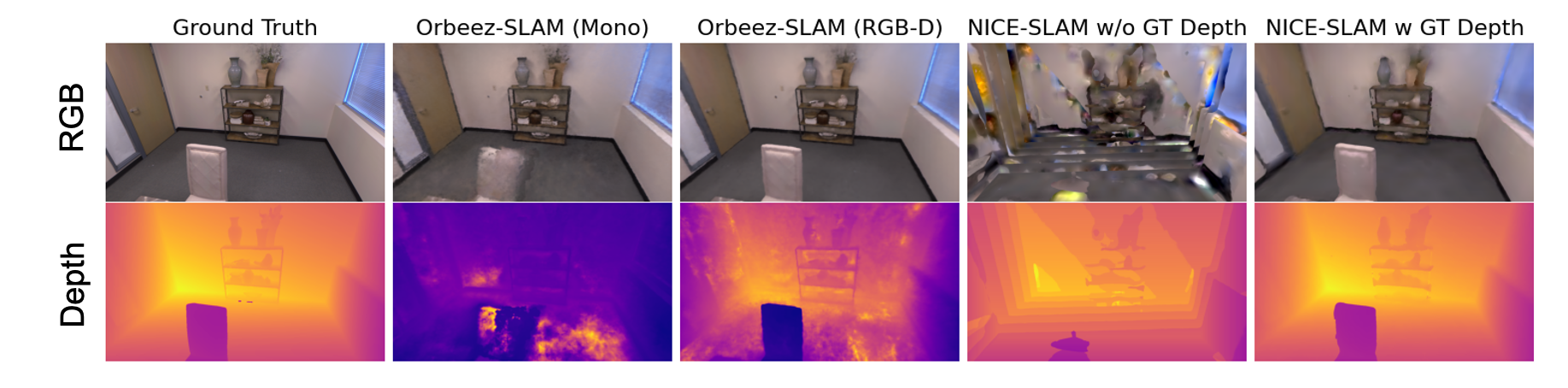}
    \caption{\textbf{Comparison of Rendering Results}. RGB and Depth of NeRF-rendered results from Orbeez-SLAM (ours) and NICE-SLAM \cite{Zhu2022CVPR} are visualized. We provide the results of Orbeez-SLAM (mono and RGB-D) and NICE-SLAM (RGB-D w/ and w/o GT depth during inference). Notably, we do not use depth information for NeRF rendering in the RGB-D setting (depth is only used for the tracking process); thus, NICE-SLAM provides a better depth rendering result. }
    \label{fig:render}
\end{figure*}

\begin{figure}[!ht]
    \centering
    \includegraphics[width=\linewidth]{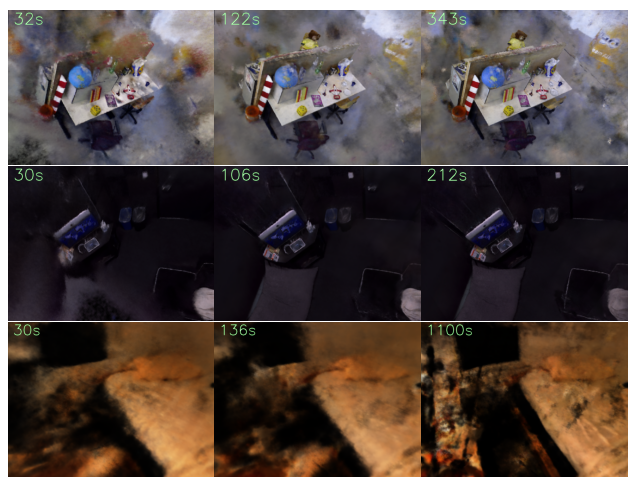}
    \caption{\textbf{NeRF Results across Time}. Our NeRF-rendered results from TUM-fr3/office, Replica-office, and ScanNet-0207 are listed across time. The elapsed time is indicated at the top left corner. The first, second, and third columns are the rendering results at the beginning of training, the end of the tracking process, and the full convergence of loss values, respectively. Our NeRF computes good results on TUM and Replica but fails on ScanNet (large scene). It successfully reconstructs the bed but fails to rebuild the desk on the left, revealing that large scenes are more challenging.}
    \label{fig:3x3}
\end{figure}

\subsubsection{Evaluation on Replica \cite{replica19arxiv}.} NICE-SLAM evaluates the mapping results on Replica since it provides GT meshes. But, as stated before, we argue that the mesh generation process from NeRF is not unified and tricky. Hence, we use common metrics in NeRF works, Depth L1 and PSNR.

As demonstrated in TABLE \ref{tab:replica_rec}, NICE-SLAM obtains the best values on the Depth L1 when GT depth is supported during rendering depth. However, our Depth L1 values outperform NICE-SLAM when it has no GT depth in rendering. Note that our NeRF is never supervised by GT depth. Next, when comparing the quality of rendered images from NeRF, we beat all variants of NICE-SLAM on PSNR, indicating that our method provides a superior color result.
\subsubsection{Runtime Comparison.} TABLE \ref{tab:speed} depicts the elapsed time of our Orbeez-SLAM and NICE-SLAM running on the TUM RGB-D benchmark. Attributed to the VO for estimating an accurate camera pose at the early stage of training, Orbeez-SLAM is 360 $\sim$ 800 times faster than NICE-SLAM.

\subsection{Ablation Study} 
TABLE \ref{tab:ablation} illustrates the ablations. We can observe that the camera pose guided only by $L_{rpj}$ achieves a better result than the one guided by both $L_{rpj}$ and $L_{pht}$ (from NeRF). We claim that the convergence speed of $L_{pht}$ is much slower, which brings a negative influence when being leveraged in real-time inference. And that is also the reason we did not provide the version only guided by $L_{pht}$ since it produces horrible results and is not available under our setting. We refer interested readers to our demo video for more details.


\subsection{Qualitative Results}
We deliver qualitative results in Fig. \ref{fig:render} and Fig. \ref{fig:3x3}. As stated in Fig. \ref{fig:render}, NICE-SLAM renders images with the help of GT depth. To be clear, GT depth is used during training of both NICE-SLAM cases. By contrast, our Orbeez-SLAM does not use depth supervision to render images, even in the RGB-D case where the GT depth is only used for tracking. Notably, Orbeez-SLAM provides a superior RGB result than NICE-SLAM under both settings. We highlight that NICE-SLAM produces better depth results due to accessing GT depth. 

Besides, we provide Orbeez-SLAM rendered results at distinct timestamps in Fig \ref{fig:3x3}. After the real-time SLAM is ended (second column), we apply offline training for NeRF until losses fully converge (third column). Orbeez-SLAM demonstrates excellent outcomes in TUM and Replica cases (first two rows) but fails at the large-scale ScanNet case. We assume large-scale scenes are more challenging to Orbeez-SLAM, and we leave it as one of future works. 

\section{Conclusion}
We aim to develop a core component in spatial AI, i.e., a pre-training-free visual SLAM that reaches real-time inference and provides dense maps for downstream tasks. To this end, we propose Orbeez-SLAM, which utilizes ORB features and NeRF-realized mapping. We cooperate with visual odometry and fast NeRF implementation on the instant-ngp platform. Moreover, Orbeez-SLAM can work with monocular cameras, leading to flexible, practical applications. We believe we pave the way for speeding up the development progress of spatial AI. Notably, how to effectively leverage dense maps in downstream tasks is interesting but is out of the scope of this paper; we leave it as future works.


\section*{Acknowledgement}
This work was supported in part by the National Science and Technology Council, under Grant MOST 110-2634-F-002-051, Qualcomm through a Taiwan University Research Collaboration Project, Mobile Drive Technology Co., Ltd (MobileDrive), and NOVATEK fellowship. We are grateful to the National Center for High-performance Computing.











\clearpage
\bibliographystyle{IEEEtran}
\bibliography{reference}

\begin{thebibliography}{10}
\providecommand{\url}[1]{#1}
\csname url@samestyle\endcsname
\providecommand{\newblock}{\relax}
\providecommand{\bibinfo}[2]{#2}
\providecommand{\BIBentrySTDinterwordspacing}{\spaceskip=0pt\relax}
\providecommand{\BIBentryALTinterwordstretchfactor}{4}
\providecommand{\BIBentryALTinterwordspacing}{\spaceskip=\fontdimen2\font plus
\BIBentryALTinterwordstretchfactor\fontdimen3\font minus
  \fontdimen4\font\relax}
\providecommand{\BIBforeignlanguage}[2]{{%
\expandafter\ifx\csname l@#1\endcsname\relax
\typeout{** WARNING: IEEEtran.bst: No hyphenation pattern has been}%
\typeout{** loaded for the language `#1'. Using the pattern for}%
\typeout{** the default language instead.}%
\else
\language=\csname l@#1\endcsname
\fi
#2}}
\providecommand{\BIBdecl}{\relax}
\BIBdecl

\bibitem{murORB2}
R.~Mur-Artal and J.~D. Tard\'os, ``{ORB-SLAM2}: an open-source {SLAM} system
  for monocular, stereo and {RGB-D} cameras,'' \emph{IEEE Transactions on
  Robotics}, vol.~33, no.~5, pp. 1255--1262, 2017.

\bibitem{engel14eccv}
J.~Engel, T.~Schöps, and D.~Cremers, ``{LSD-SLAM}: Large-scale direct
  monocular {SLAM},'' in \emph{European Conference on Computer Vision (ECCV)},
  September 2014.

\bibitem{koestler2021tandem}
L.~Koestler, N.~Yang, N.~Zeller, and D.~Cremers, ``Tandem: Tracking and dense
  mapping in real-time using deep multi-view stereo,'' in \emph{Conference on
  Robot Learning (CoRL)}, 2021.

\bibitem{mildenhall2020nerf}
B.~Mildenhall, P.~P. Srinivasan, M.~Tancik, J.~T. Barron, R.~Ramamoorthi, and
  R.~Ng, ``Nerf: Representing scenes as neural radiance fields for view
  synthesis,'' in \emph{ECCV}, 2020.

\bibitem{Sucar:etal:ICCV2021}
E.~Sucar, S.~Liu, J.~Ortiz, and A.~Davison, ``{iMAP}: Implicit mapping and
  positioning in real-time,'' in \emph{Proceedings of the International
  Conference on Computer Vision ({ICCV})}, 2021.

\bibitem{Zhu2022CVPR}
Z.~Zhu, S.~Peng, V.~Larsson, W.~Xu, H.~Bao, Z.~Cui, M.~R. Oswald, and
  M.~Pollefeys, ``Nice-slam: Neural implicit scalable encoding for slam,'' in
  \emph{Proceedings of the IEEE/CVF Conference on Computer Vision and Pattern
  Recognition (CVPR)}, 2022.

\bibitem{mueller2022instant}
\BIBentryALTinterwordspacing
T.~M\"uller, A.~Evans, C.~Schied, and A.~Keller, ``Instant neural graphics
  primitives with a multiresolution hash encoding,'' \emph{ACM Trans. Graph.},
  vol.~41, no.~4, pp. 102:1--102:15, Jul. 2022. [Online]. Available:
  \url{https://doi.org/10.1145/3528223.3530127}
\BIBentrySTDinterwordspacing

\bibitem{tiny-cuda-nn}
T.~M\"uller, ``Tiny {CUDA} neural network framework,'' 2021,
  https://github.com/nvlabs/tiny-cuda-nn.

\bibitem{takikawa2021nglod}
T.~Takikawa, J.~Litalien, K.~Yin, K.~Kreis, C.~Loop, D.~Nowrouzezahrai,
  A.~Jacobson, M.~McGuire, and S.~Fidler, ``Neural geometric level of detail:
  Real-time rendering with implicit {3D} shapes,'' 2021.

\bibitem{Park_2019_CVPR}
J.~J. Park, P.~Florence, J.~Straub, R.~Newcombe, and S.~Lovegrove, ``Deepsdf:
  Learning continuous signed distance functions for shape representation,'' in
  \emph{The IEEE Conference on Computer Vision and Pattern Recognition (CVPR)},
  June 2019.

\bibitem{martinbrualla2020nerfw}
R.~Martin-Brualla, N.~Radwan, M.~S.~M. Sajjadi, J.~T. Barron, A.~Dosovitskiy,
  and D.~Duckworth, ``{NeRF in the Wild: Neural Radiance Fields for
  Unconstrained Photo Collections},'' in \emph{CVPR}, 2021.

\bibitem{barron2021mipnerf}
J.~T. Barron, B.~Mildenhall, M.~Tancik, P.~Hedman, R.~Martin-Brualla, and P.~P.
  Srinivasan, ``Mip-nerf: A multiscale representation for anti-aliasing neural
  radiance fields,'' \emph{ICCV}, 2021.

\bibitem{barron2022mipnerf360}
J.~T. Barron, B.~Mildenhall, D.~Verbin, P.~P. Srinivasan, and P.~Hedman,
  ``Mip-nerf 360: Unbounded anti-aliased neural radiance fields,'' \emph{CVPR},
  2022.

\bibitem{schoenberger2016sfm}
J.~L. Sch\"{o}nberger and J.-M. Frahm, ``Structure-from-motion revisited,'' in
  \emph{Conference on Computer Vision and Pattern Recognition (CVPR)}, 2016.

\bibitem{schoenberger2016mvs}
J.~L. Sch\"{o}nberger, E.~Zheng, M.~Pollefeys, and J.-M. Frahm, ``Pixelwise
  view selection for unstructured multi-view stereo,'' in \emph{European
  Conference on Computer Vision (ECCV)}, 2016.

\bibitem{wang2021nerfmm}
Z.~Wang, S.~Wu, W.~Xie, M.~Chen, and V.~A. Prisacariu, ``Ne{RF}$--$: Neural
  radiance fields without known camera parameters,'' \emph{arXiv preprint
  arXiv:2102.07064}, 2021.

\bibitem{SCNeRF2021}
Y.~Jeong, S.~Ahn, C.~Choy, A.~Anandkumar, M.~Cho, and J.~Park,
  ``Self-calibrating neural radiance fields,'' in \emph{Proceedings of the
  IEEE/CVF International Conference on Computer Vision (ICCV)}, October 2021,
  pp. 5846--5854.

\bibitem{lin2021barf}
C.-H. Lin, W.-C. Ma, A.~Torralba, and S.~Lucey, ``Barf: Bundle-adjusting neural
  radiance fields,'' in \emph{IEEE International Conference on Computer Vision
  ({ICCV})}, 2021.

\bibitem{6162880}
R.~A. Newcombe, S.~Izadi, O.~Hilliges, D.~Molyneaux, D.~Kim, A.~J. Davison,
  P.~Kohi, J.~Shotton, S.~Hodges, and A.~Fitzgibbon, ``Kinectfusion: Real-time
  dense surface mapping and tracking,'' in \emph{2011 10th IEEE International
  Symposium on Mixed and Augmented Reality}, 2011, pp. 127--136.

\bibitem{8255617}
E.~Vespa, N.~Nikolov, M.~Grimm, L.~Nardi, P.~H.~J. Kelly, and S.~Leutenegger,
  ``Efficient octree-based volumetric slam supporting signed-distance and
  occupancy mapping,'' \emph{IEEE Robotics and Automation Letters}, vol.~3,
  no.~2, pp. 1144--1151, 2018.

\bibitem{murAcceptedTRO2015}
R.~Mur-Artal, J.~M.~M. Montiel, and J.~D. Tardós, ``Orb-slam: A versatile and
  accurate monocular slam system,'' \emph{IEEE Transactions on Robotics},
  vol.~31, no.~5, pp. 1147--1163, 2015.

\bibitem{ORBSLAM3_TRO}
C.~Campos, R.~Elvira, J.~J.~G. Rodríguez, J.~M. M.~Montiel, and J.~D.~Tardós,
  ``Orb-slam3: An accurate open-source library for visual, visual–inertial,
  and multimap slam,'' \emph{IEEE Transactions on Robotics}, vol.~37, no.~6,
  pp. 1874--1890, 2021.

\bibitem{468400}
N.~Max, ``Optical models for direct volume rendering,'' \emph{IEEE Transactions
  on Visualization and Computer Graphics}, vol.~1, no.~2, pp. 99--108, 1995.

\bibitem{liu2020neural}
L.~Liu, J.~Gu, K.~Z. Lin, T.-S. Chua, and C.~Theobalt, ``Neural sparse voxel
  fields,'' \emph{NeurIPS}, 2020.

\bibitem{yu_and_fridovichkeil2021plenoxels}
{Sara Fridovich-Keil and Alex Yu}, M.~Tancik, Q.~Chen, B.~Recht, and
  A.~Kanazawa, ``Plenoxels: Radiance fields without neural networks,'' in
  \emph{CVPR}, 2022.

\bibitem{SunSC22}
C.~Sun, M.~Sun, and H.~Chen, ``Direct voxel grid optimization: Super-fast
  convergence for radiance fields reconstruction,'' in \emph{CVPR}, 2022.

\bibitem{Clark_2022_CVPR}
R.~Clark, ``Volumetric bundle adjustment for online photorealistic scene
  capture,'' in \emph{Proceedings of the IEEE/CVF Conference on Computer Vision
  and Pattern Recognition (CVPR)}, June 2022, pp. 6124--6132.

\bibitem{DBLP:journals/corr/abs-1912-03858}
\BIBentryALTinterwordspacing
Y.~Chen, Y.~Chen, and G.~Wang, ``Bundle adjustment revisited,'' \emph{CoRR},
  vol. abs/1912.03858, 2019. [Online]. Available:
  \url{http://arxiv.org/abs/1912.03858}
\BIBentrySTDinterwordspacing

\bibitem{sturm12iros}
J.~Sturm, N.~Engelhard, F.~Endres, W.~Burgard, and D.~Cremers, ``A benchmark
  for the evaluation of rgb-d slam systems,'' in \emph{Proc. of the
  International Conference on Intelligent Robot Systems (IROS)}, Oct. 2012.

\bibitem{replica19arxiv}
J.~Straub, T.~Whelan, L.~Ma, Y.~Chen, E.~Wijmans, S.~Green, J.~J. Engel,
  R.~Mur-Artal, C.~Ren, S.~Verma, A.~Clarkson, M.~Yan, B.~Budge, Y.~Yan,
  X.~Pan, J.~Yon, Y.~Zou, K.~Leon, N.~Carter, J.~Briales, T.~Gillingham,
  E.~Mueggler, L.~Pesqueira, M.~Savva, D.~Batra, H.~M. Strasdat, R.~D. Nardi,
  M.~Goesele, S.~Lovegrove, and R.~Newcombe, ``The {R}eplica dataset: A digital
  replica of indoor spaces,'' \emph{arXiv preprint arXiv:1906.05797}, 2019.

\bibitem{dai2017scannet}
A.~Dai, A.~X. Chang, M.~Savva, M.~Halber, T.~Funkhouser, and M.~Nie{\ss}ner,
  ``Scannet: Richly-annotated 3d reconstructions of indoor scenes,'' in
  \emph{Proc. Computer Vision and Pattern Recognition (CVPR), IEEE}, 2017.

\bibitem{huang2021difusion}
J.~Huang, S.-S. Huang, H.~Song, and S.-M. Hu, ``Di-fusion: Online implicit 3d
  reconstruction with deep priors,'' in \emph{Proceedings of the IEEE/CVF
  Conference on Computer Vision and Pattern Recognition}, 2021.

\bibitem{8954208}
T.~Schöps, T.~Sattler, and M.~Pollefeys, ``Bad slam: Bundle adjusted direct
  rgb-d slam,'' in \emph{2019 IEEE/CVF Conference on Computer Vision and
  Pattern Recognition (CVPR)}, 2019, pp. 134--144.

\bibitem{whelan2012kintinuous}
T.~Whelan, M.~Kaess, M.~Fallon, H.~Johannsson, J.~Leonard, and J.~McDonald,
  ``Kintinuous: Spatially extended kinectfusion,'' 2012.

\bibitem{5979949}
R.~Kümmerle, G.~Grisetti, H.~Strasdat, K.~Konolige, and W.~Burgard, ``G2o: A
  general framework for graph optimization,'' in \emph{2011 IEEE International
  Conference on Robotics and Automation}, 2011, pp. 3607--3613.

\end{thebibliography}

\end{document}